\documentclass[conference]{IEEEtran}
\IEEEoverridecommandlockouts
\usepackage{cite}
\usepackage{amsmath,amssymb,amsfonts}
\usepackage{algorithmic}
\usepackage{graphicx}
\usepackage{textcomp}
\usepackage{xcolor}
\usepackage{hyperref}
\usepackage{caption}
\usepackage{subcaption}

\def\BibTeX{{\rm B\kern-.05em{\sc i\kern-.025em b}\kern-.08em
    T\kern-.1667em\lower.7ex\hbox{E}\kern-.125emX}}
\begin{document}

\title{BLUE: A 3D Dynamic Bipedal Robot}

\author{\IEEEauthorblockN{Germán David Vargas Gutiérrez}
\IEEEauthorblockA{
\textit{Mechatronics Engineering}\\
Universidad de San Buenaventura,\\
Bogot\'a D.C., Colombia,\\ 
gdvargas@academia.usbbog.edu.co}
\and
\IEEEauthorblockN{Margarita Rosa Berrio Soto}
\IEEEauthorblockA{
\textit{Mechatronics Engineering}\\
Universidad de San Buenaventura,\\
Bogot\'a D.C., Colombia,\\ 
mberrio@academia.usbbog.edu.co}
\and
\IEEEauthorblockN{Jaime Arcos-Legarda}
\IEEEauthorblockA{\textit{Full professor Mechatronics Engineering} \\
Universidad de San Buenaventura,\\
Bogot\'a D.C., Colombia,\\ 
warcos@usbbog.edu.co}
}

\maketitle

\begin{abstract}
The objective of this work is to design a mechatronic bipedal robot with mobility in 3D environments. The designed robot has a total of six actuated degrees of freedom (DoF), each leg has two DoF located at the hip: one for abduction/adduction and another for thigh flexion/extension, and a third DoF at the knee for the shin flexion/extension. This robot is designed with point-feet legs to achieve a dynamic underactuated walking. Each actuator in the robot includes a DC gear motor, an encoder for position measurement, a flexible joint to form a series flexible actuator, and a feedback controller to ensure trajectory tracking. In order to reduce the total mass of the robot, the shin is designed using topology optimization. The resulting design is fabricated using 3D printed parts, which allows to get a robot's prototype to validate the selection of actuators. The preliminary experiments confirm the robot's ability to maintain a stand-up position and let us drawn future works in dynamic control and trajectory generation for periodic stable walking.
\end{abstract}


\section{Introduction}

At the end of the 20th century, Quiang et al., developed a method to plan a biped gait of a bipedal robot dividing the task into a foot and hip trajectories  \cite{huang1999high}. This was carried out through a gait pattern generator, which used constraints generated with a third order spline interpolation. These constraints were imposed on the foot and  hip trajectories. Later on, at the beginning of the 21st century, Gienger et al., designed an anthropomorphic bipedal robot to achieve a dynamic three-dimensional gait using weight reduction methods \cite{gienger2001towards}.
Weng specifically focused on modeling the three-dimensional dynamics of a humanoid robot, known as TEO, using Webots as a simulator, where, basically, its purpose was to show the robot's movement in a virtual environment to perform predefined tasks  \cite{weng2012creacion}. For its development, it started with the creation of eight nodes, which contemplated the number of joints that are connected, then they analyzed the degrees of freedom existing in the mechanism. Using an equivalent approach, Park et al., \cite{park2010identification} identify the dynamic model of a bipedal robot, known as MABEL, which includes a set of differential cables and integrates a torso assembled with two legs, in which each one functions as a monopod with a transmission mechanism implemented with the differential cable. An evolution of MABEL was developed with the 3D bipedal robot ATRIAS \cite{hubicki2016atrias}.

\vspace{2mm}
%on the other hand \cite{arcos2019robust} A control strategy is proposed for application in the stable gait of bipedal robots with random perturbations present in the trajectory. For this, a trajectory generator is applied that defines the parameters to be followed by each of the actuators present in the leg, mainly for the change of the leg during the walk. For the implementation, use is made of a generalized multivariable PI controller without switching, which is in charge of rejecting the disturbances present in the environment to which the robot is running. Finally, this control strategy was validated with the application to a numerical model of a five-link bipedal robot, where the results demonstrated its efficiency with good performance and robustness. 

%\vspace{3mm}
%In \cite{arcos2019hybrid} The use of continuous-time control strategies is considered and discreet. In this way, the rejection of the disturbances is made by tracking the references during the trajectory, while at the same time the trajectory references are reestablished after each impact generated in the change of support foot during the walk, thus guaranteeing zero error in tracking.
On the other hand, \cite{arcos2019robust} A control strategy is proposed for its application in the stable gait of bipedal robots with random disturbances present in the trajectory. To do this, a trajectory generator is applied that defines the parameters to be followed by each of the actuators present in the leg, mainly for the change of leg during the walk. For the implementation, use is made of a generalized multivariable PI controller without switching, which is responsible for rejecting the disturbances present in the environment to which the robot is directed. Some time later, in \cite{arcos2019hybrid}, the development of control techniques in continuous and discrete time continued with a focus on following the trajectories for each joint and where the impact generated on the leg by the change of support in the walk was taken into account.

\vspace{2mm}
In \cite{hurst2008role} the most common failure with respect to the dynamic gait of bipedal robots is evidenced, and it is that they highlight that the limitations in performance are mainly due to the limitations caused by the mechanical design since to date most were built with complex mechanisms that hindered the transmission of kinetic energy and the focus on rigid transmissions of motion. That is why the use of springs as an improvement for the kinematic design of the legs is proposed as a solution, achieving a better energy storage, and a high mechanical power transmission. In the same vein, the ATRIAS 2.1 robot is developed in search of efficiency in the transfer of energy, speed and above all robustness with respect to the terrain of the walk, which seeks not to depend on the vision of the robot. Once raised, as indicated in \cite{ramezani2014performance}, the legs are designed to be light, specifically less than 5 percent of the total mass, which are actuated by series-connected spring actuators. Like its predecessors, the ATRIAS 2.1 robot makes use of the inverted pendulum mechanism.

\vspace{1mm}
A more recent design is the one presented in \cite{hattori2020design}, where, at MIT, they made a Mini Cheetah where one of its key peculiarities is the redesign of the actuators and joints since the components are easy to replace due to the size and simpler. mechanics. In the document they focus on the design and manufacture of Mini Cheetah actuators, which correspond to a new generation of these. Modular actuators have the particularity that they have a brushless motor and the gearbox is stationary, this differs from the original because the motor was in series with the actuator.

\section{Design requirements}
The design requirements of the BLUE bipedal robot are established, which were to generate a structure from the use of independent mechanisms that offered the option of being modular with an operating torque that did not exceed 2 Nm. In addition to this, the desired average angular velocity for the mobility of each section was taken into account, where a value of 70 RPM was validated at the output of each of the actuators.

\section{Design of bipedal robot BLUE}

\begin{figure}[ht]
	\centering
		\includegraphics[trim=0 0 0 0,clip,width=0.48\textwidth]{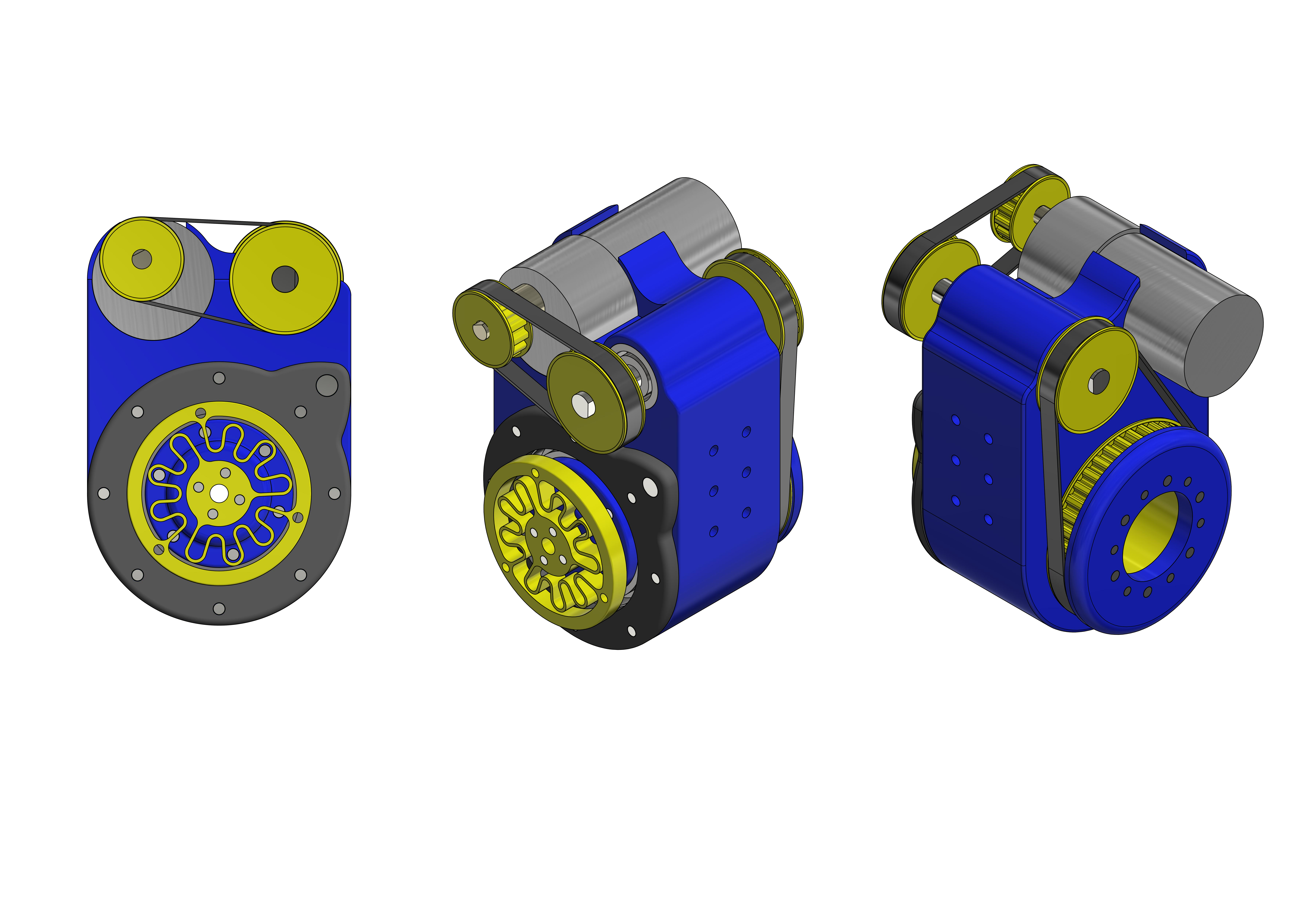}
		\caption{Mechanical actuator}
	\label{fig:MechanicalActuator}
\end{figure}

The development of the BLUE bipedal robot is based on the use of six (6) mechanical actuators, each strategically located to simulate the three (3) degrees of freedom present in the proposed leg, these have a DC motor with a gearbox in such a way that after the implementation of some logic module such as H bridges it is possible to control the direction of rotation of each body. The representation is evidenced in the \autoref{fig:MechanicalActuator}

The structure of the mechanical actuator is based on the principle of reduction by use of synchronous belts where the first pulley, assembled to the motor shaft, transmits the movement by means of a 70XL reference toothed belt to a larger diameter pulley coupled to a transmission shaft, which at in turn, it transmits the energy through a 116XL reference toothed belt, with which in the end a reduction was obtained at the actuator output, in addition, it is worth highlighting one of the main strengths of the actuator design, which is to be a modular device.

\begin{table}[ht]
\caption{Synchronous belts for transmission}
\begin{center}
\begin{tabular}{| c | c | c | c | c |}
\hline
\multicolumn{5}{ |c| }{Synchronouss Belts Powergrip} \\ \hline
Description &   Pitch length  & Number of  &   Step   & Belt height \\
&  [mm] & teeth  &  [mm]   &  [mm] \\\hline
70XL  & 177.80 & 35 & 5.08 & 2.3 \\
116XL & 294.64 & 58 & 5.08 & 2.3 \\ \hline
\end{tabular}
\label{tab:CorreasSincronas}
\end{center}
\end{table}

From the belt transmission system generated with the help of the Inventor software, the calculation of the angular velocities and the torques present along the axes was carried out and evidence was left in the \autoref{tab:VelocidadesyTorquesIO}. Which was completed as of \autoref{eq:VelocidadAngularSalida}, \autoref{eq:velocidadAngularEje}, \autoref{eq:TorqueEje} and \autoref{eq:TorqueSalida}

\begin{equation}
    \omega_{out}=\frac{D_{p1} D_{p3} \omega_{int}}{D_{p2} D_{p4}}
    \label{eq:VelocidadAngularSalida}
\end{equation}

\begin{equation}
    \omega_{axle}=\frac{D_{p1} \omega_{int}}{D_{p2}}
    \label{eq:velocidadAngularEje}
\end{equation}

\begin{equation}
    \tau_{axle}=\frac{P_m}{\omega_{axle}}
    \label{eq:TorqueEje}
\end{equation}

\begin{equation}
    \tau_{out}=\frac{P}{\omega_{out}}
    \label{eq:TorqueSalida}
\end{equation}

where:

\begin{center}
$D_p$ $\Longrightarrow$ Pitch diameter's

$\omega_{out}$ $\Longrightarrow$ Output angular velocity

$\omega_{axle}$ $\Longrightarrow$ axle angular velocity

$\tau_{out}$ $\Longrightarrow$ Output torque

$\tau_{axle}$ $\Longrightarrow$ Axle torque
\end{center}

\begin{table}[ht]
\caption{Angular velocity and torque}
\begin{center}
\begin{tabular}{| c | c | c | c | c | c |}
\hline

 \multicolumn{3}{ |c|}{Angular velocity [rpm]} &  \multicolumn{3}{ |c|}{Torque [Nm]}\\ \hline
$\omega_{int}$ &  $\omega_{out}$   & $\omega_{axle}$   & $\tau_{int}$  & $\tau_{out}$ & $\tau_{axle}$ \\ \hline
29 & 11.48 & 21.75 & 2.63 & 6.66 & 3.51\\
37 & 14.57 & 27.75 & 2.06 & 5.24 & 2.75\\
45 & 17.72 & 33.75 & 1.69 & 4.31 & 2.26\\
51 & 20.08 & 38.25 & 1.49 & 3.80 & 1.99\\
67 & 26.38 & 50.25 & 1.14 & 2.89 & 1.52\\ 
74 & 29.14 & 55.5  & 1.03 & 2.62 & 1.37\\ 
81 & 31.89 & 60.75 & 0.94 & 2.40 & 1.26\\
98 & 38.58 & 73.5  & 0.78 & 1.98 & 1.04\\
107 & 42.13 & 80.25 & 0.71 & 1.81 & 0.95\\
121 & 47.64 & 90.75 & 0.63 & 1.60 & 0.84\\ \hline
\end{tabular}
\label{tab:VelocidadesyTorquesIO}
\end{center}
\end{table}

\begin{itemize}
    \item Transmission axis design

%     \begin{figure}[ht]
% 	\centering
% 		\includegraphics[trim=0 0 0 0,clip,width=0.40\textwidth]{EJE TRANSMISOR.JPG}
% 		\caption{transmission shaft}
% 	\label{fig:transmission shaft}
%     \end{figure}
    \begin{figure}[ht]
	\centering
		\includegraphics[trim=0 0 0 0,clip,width=0.40\textwidth]{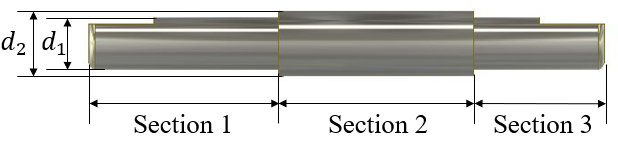}
		\caption{Transmission shaft}
	\label{fig:transmission shaft}
    \end{figure}

    The transmission shaft was machined in silver steel as a greater resistance to the stress that would arise in the use of the composite belt system was required. For the stress analysis, the ASME - elliptical theory mentioned in \autoref{eq:asme eliptica} and transformed according to the forces present in each section was used.

    \begin{equation}
   \left ( \frac{S_a}{S_e} \right )^2 +  \left ( \frac{S_m}{S_y} \right )^2 = 1
   \label{eq:asme eliptica}
    \end{equation}

    \vspace{0.5mm}
    After the stress analysis, the minimum diameters specified in the design were validated, which were $ d_1=6 [mm] $, $ d_2=7.6 [mm] $, and $ d_1=d_3 $ this because they were capable of withstand torques calculated in \autoref{tab:VelocidadesyTorquesIO}
    
    %the minimum diameters%
    
    \vspace{2mm}
    \item Internal axis design
    
    In the same way, for the design of the internal axis evidenced in \autoref{fig:Internal_shaft}, the ASME - elliptical theory mentioned in \autoref{eq:asme eliptica} was taken into account, however, for this case it was verified that in none of the different sections along the axis the calculated torque was exceeded.
    
    \begin{equation}
        T_a = \sqrt{\frac{1}{768 k_{ft}}\left( \frac{\pi^2 \left(D^4 - d^4 \right)^2 S_e ^2}{fs^2 D^2} - \left( 32k_{ff}M_a \right)^2 \right)}
        \label{eq:TorqueAlternanteSec2}
    \end{equation}
    
    \vspace{0.5mm}
    It was determined that along the axis, the lowest torque capable of withstanding was $ \ tau = 40 [Nm] $ value that when compared with \autoref{tab:VelocidadesyTorquesIO} guarantees that at no time the axis would fail due to stress cutting. In this way, the design of the shaft was validated, which should be noted that it was made up of pieces printed in PLA material with a 90-percent filling and rigidly joined.
    
    \vspace{1mm}
    \item Base design
    \begin{figure}[ht]
	\centering
    \begin{subfigure}[b]{0.24\textwidth}
         \centering
         \includegraphics[width=\textwidth]{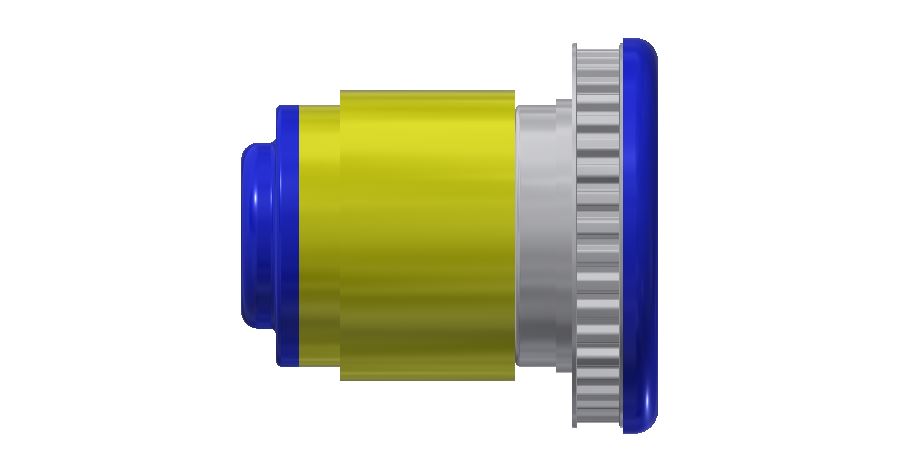}
		\caption{}
	    \label{fig:Internal_shaft}
     \end{subfigure}
          %\hfill
     \begin{subfigure}[b]{0.24\textwidth}
         \centering
         \includegraphics[width=\textwidth]{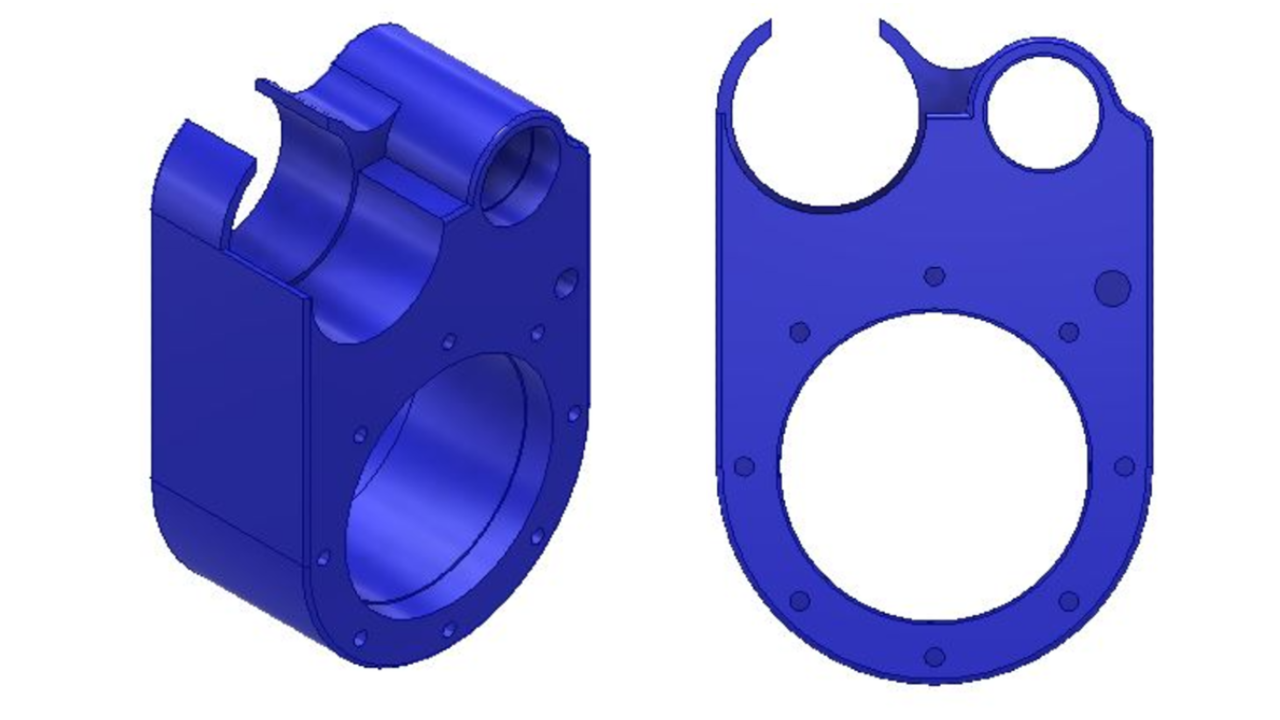}
		\caption{}
	    \label{fig:basedesign}
     \end{subfigure}
     \caption{a) Internal shaft. b) Actuator base}
    \label{fig:three_graphs}
    \end{figure}
    
    The base of the mechanical actuator was one of the principal pieces to take into account within the design due to its great effort to contain the system made up of belts and the need for a structure capable of protecting the position of the motor and the eventual vibrations that would come to throughout the experimental tests. The critical points of the structure evidenced in \autoref{fig:basedesign} were analyzed, and the maximum force capable of withstanding was determined for each case.
    
    \begin{equation}
        \frac{\sigma_a^2+3\tau_a^2}{S_e^2}=\frac{1}{f.s^2}
        \label{eq:inicialTotal}
    \end{equation}
    
    \vspace{1mm}
    For each case, the corresponding efforts were replaced together in \autoref{eq:inicialTotal} with the known data of the torques exerted in each of the sections, where it was determined that the critical zone in the event of possible failure was the motor support with a force $ F = 8 [ N] $ it is evidenced in \autoref{eq:basefinal}, however, as it is not a more demanding part, it was validated and implemented in the final design.
    
    \begin{equation}
        \frac{1}{\left(64.93\ Mpa\right)^2}\left(\frac{\left(F\right)\left(1.57\right)}{5.645x{10}^{-7}\left[m\right]}\left(1.45\right)\right)^2=\frac{1}{4}
        \label{eq:basefinal}
    \end{equation}
    
    \[F=8.15N\]
    
    \item Thigh Design
    
    \begin{figure}[ht]
	\centering
		\includegraphics[trim=0 0 0 0,clip,width=0.45\textwidth]{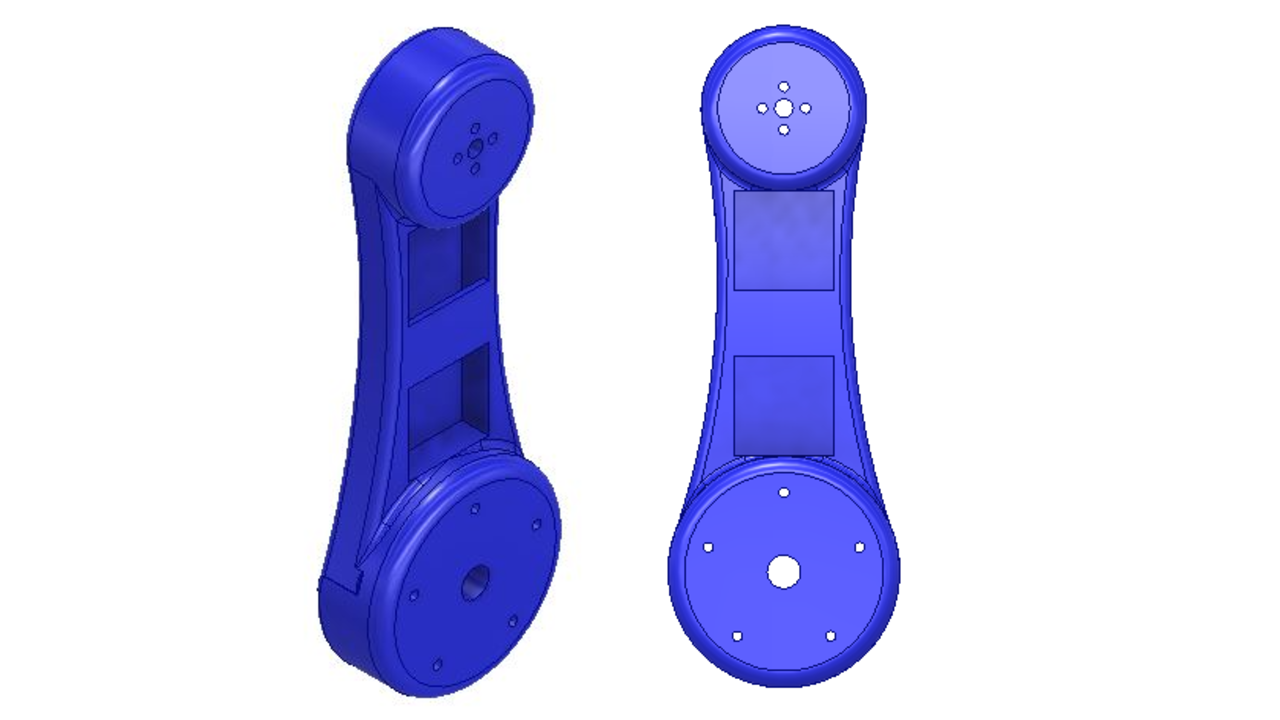}
		\caption{Femur of bipedal robot BLUE}
	\label{fig:femurdesign}
    \end{figure}
    
    The femur of the BLUE robot was designed from the requirement of the interconnection between the actuator located in the thigh and the actuator of the knee, where, in addition to maintaining the limit in the standard measures proposed for the robot, it was defined as a restriction the internal location of the actuators, which also allowed the correct distribution of the electrical wiring along the link, as observed in \autoref{fig:femurdesign}
    
    \begin{equation}
        \sigma=\frac{13N}{(A)}+\frac{(M)(c)}{I}
        \label{eq:femurprincipal}
    \end{equation}
    
    Because the femur is a piece subjected to bending and tension, the stress analysis was carried out to determine the maximum force capable of withstanding, for which the data provided by Inventor were taken and to have a range of action in the tensile force, the which was approximately $ 13 [N] $ as evidenced in \autoref{eq:femurbien}
    
    \begin{equation}
        \frac{1}{\left(60Mpa\right)^2}\left(\frac{\left(57.5N\right)}{\left(2.4x{10}^{-4}m\right)}+\frac{\left(F\right)\left(190x{10}^{-5}\right)}{7.246x{10}^{-9}}\right)^2=\frac{1}{4}
    \label{eq:femurbien}
    \end{equation}
    
    \[F=113.80N\]
    
    \vspace{1mm}
    It was determined that the change in the lower section of the piece was the most prone to failure, although with a fairly high operating range, since a maximum bending force of $ 110 [N] $ was selected. To reduce risks in the experimental tests, it was proposed to make use of highlighted joints in such a way as to reduce the existing stress concentrators.
    
    \vspace{1mm}
    \item Shin Design
    
    The shin design considers a mass distribution such that the robot's dynamic model resembles an inverted pendulum model, which allows to reduce the order of the system and carry out simpler stability analysis than with the full order model.
 %  
 %  to the robot in its resting state and especially in the experimental tests, in this way the model proposed in \autoref{fig:Shindesign}. However, the amount of material for printing was considerable high, which is why a generative study of topological optimization was performed.
%     \begin{figure}[ht]
% 	\centering
% 		\includegraphics[trim=0 0 0 0,clip,width=0.45\textwidth]{shin.png}
% 		\caption{Shin of bipedal robot BLUE}
% 	\label{fig:Shindesign}
%     \end{figure}
%   
To define the shin shape a generative design is performed using topological optimization, the necessary border parameters for the operation were defined, among which it is necessary to fix the front face of the piece together with its holes with a preservation diameter of $ 8~\mbox{mm} $. Likewise, the normal tension force generated in the lower hole corresponding to $ 150~\mbox{N}$ is used, which is a data higher than the total weight of the robot in order to maintain an operating range in the physical restriction. 
The result is evidenced in \autoref{fig:Shindesignopt}
    
    \begin{figure}[ht]
	\centering
		\includegraphics[trim=0 0 0 0,clip,width=0.45\textwidth]{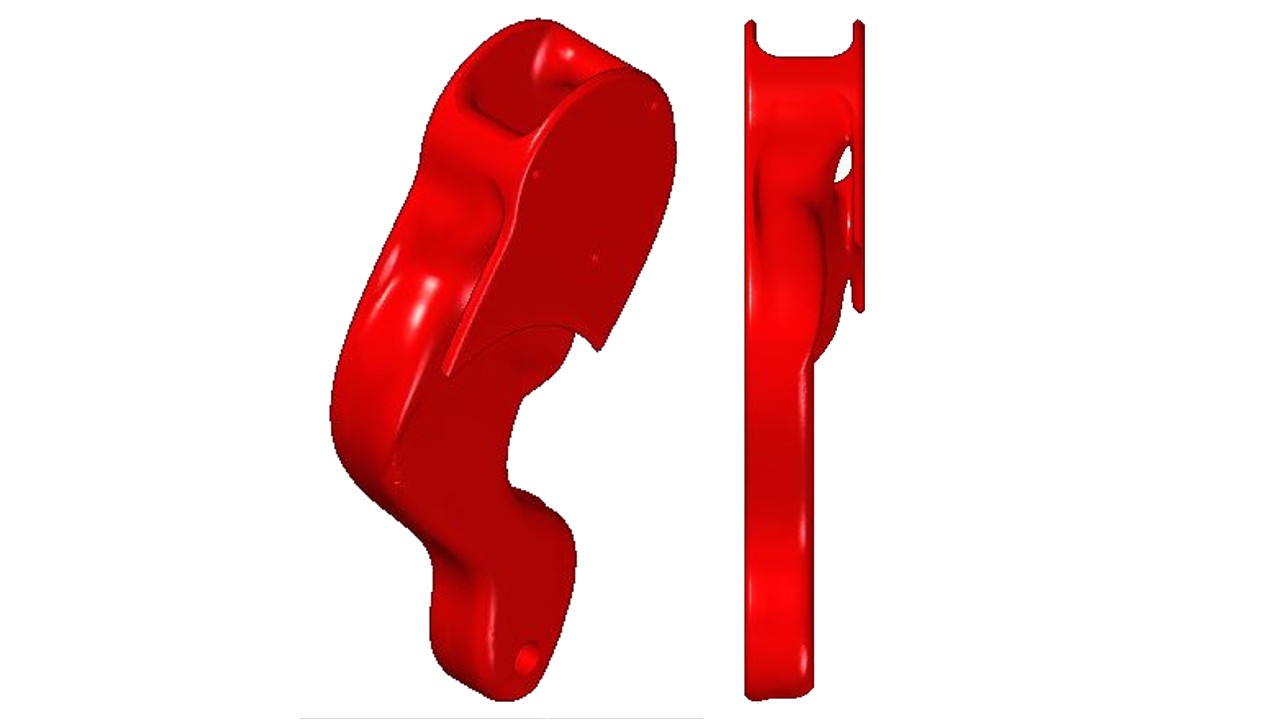}
		\caption{Shin of bipedal robot BLUE by topological optimization}
	\label{fig:Shindesignopt}
    \end{figure}
    
\end{itemize}

\begin{figure}[ht]
	\centering
		\includegraphics[trim=0 0 0 0,clip,width=0.48\textwidth]{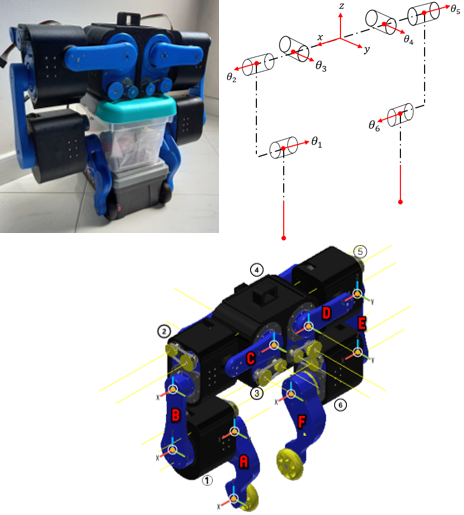}
		\caption{Bipedal robot BLUE}
	\label{fig:BLUE}
    \end{figure}

Because each of the pieces with the highest critically index in the initial prototype was validated, the preliminary assembly of the BLUE bipedal robot was carried out as evidenced in \autoref{fig:BLUE} where the names of the links and the identification of each of the mechanical actuators used.

\section{Kinematic model}
The development of the kinematic model of the bipedal robot BLUE was raised from the transformation matrices of Denavit-Hartenberg, for which the lower right-limb support was established as the point of origin, this, in order to resemble the process of locomotion in the walk to that of the human being, in which the kinematic chain begins in the ankle, passes through the hip and ends in the opposite ankle.

\begin{equation}
    \left[\begin{matrix}\begin{matrix}C(\theta_i)&-S(\theta_i)\\S\left(\theta_i\right)C(\alpha_{i-1})&C(\theta_i)C(\alpha_{i-1})\\\end{matrix}&\begin{matrix}0&a_{i-1}\\-S(\alpha_{i-1})&-S\left(\alpha_{i-1}\right)d_i\\\end{matrix}\\\begin{matrix}S(\theta_i)S(\alpha_{i-1})&C(\theta_i)S(\alpha_{i-1})\\0&0\\\end{matrix}&\begin{matrix}C\left(\alpha_{i-1}\right)\ &C\left(\alpha_{i-1}\right)d_i\\0&1\\\end{matrix}\\\end{matrix}\right]
    \label{eq:GeneralDenavit}
\end{equation}

\begin{figure}[ht]
	\centering
		\includegraphics[trim=0 0 0 0,clip,width=0.44\textwidth]{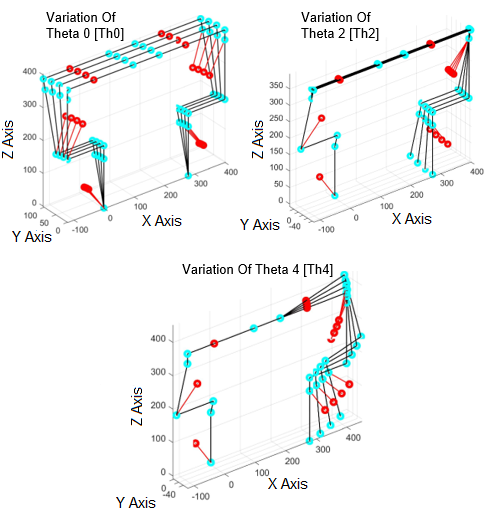}
		\caption{Variation of the angles}
	\label{fig:ANGLESTH}
\end{figure}

The kinematic model was constructed from the parameters required for the construction of the general transformation matrix described by Denavit Hartenberg evidenced in the \autoref{eq:GeneralDenavit} , in which all the frames, in addition to being in agreement with the different degrees of freedom, aligned the rotation axes (z) in such a way that along the structure, the displacement and rotation was uniform. 
\vspace{2mm}
\autoref{fig:ANGLESTH} shows variations from 0 to 15 degrees in each of the different degrees of freedom present in the BLUE kinematic chain, which are the beginning of the chain in the right support (th0), the knee of the right limb, located more specifically in the actuator (1) and one of the abductions present in the hip with the help of the actuator (4).

\newpage
\section{Dynamical model (Euler-Lagrange)}
% Six degrees of freedom are set for the Blue robot. The vector $q$ contains the angles to be varied, where $ \theta_1$ and $\theta_6$ are the angles in charge of the flexion in the knees respectively, $\theta_2$ and $\theta_5$ are the angles that perform the extension of the legs, $\theta_3$ and $\theta_4$ the ones in charge of the abduction in the hip. The location of each of the angles can be seen in each actuator, which is listed in the \autoref{fig:BLUE}.
BLUE robot have six degrees of freedom. The vector $q$ contains the angles to be varied, where $ \theta_1$ and $\theta_6$ are the angles of the knees flexion, $\theta_2$ and $\theta_5$ are the angles of the legs extension, $\theta_3$ and $\theta_4$ are the angles of the hip abduction. The location of each of the angles can be seen in the \autoref{fig:BLUE}.

Generalized coordinates are:
    
    \begin{equation}
        q=\begin{bmatrix}
       \theta_{1}&\theta_{2}&\theta_{3}&\theta_{4}&\theta_{5}&\theta_{6}
       \end{bmatrix}^T .
    \end{equation}
    
% The positions of the centers of mass of the six bodies are general define:
The positions of the centers of mass of the six bodies are:
    
    \begin{equation}
        i=\begin{bmatrix}
       X_i & Y_i & Z_i
       \end{bmatrix}^T,
       \label{eq:posicionescentrosdemasa}
    \end{equation}
       
Where $i$ varies according to the body ($A, B, C, D, E$ and $F$), $X$ is the position in x axis, $Y$ is the position in y axis and $Z$ is the position in z axis. (\autoref{eq:posicionescentrosdemasa}).

The velocity vectors ($\dot{r}$) are calculated for each body.

\[\dot{r}_i=\dot{X}_i+\dot{Y}_i+\dot{Z}_i,\]

\subsection{Calculation of the Lagrangian}

The Lagrangian is defined in the \autoref{eq:Lagrangiano}, where $T_{Total}$ is the kinetic energy and $V_{Total}$ is the potential energy.
    \begin{equation}
        \mathcal{L}:=T_{Total}-V_{Total}
        \label{eq:Lagrangiano}
    \end{equation}
    
% The angular and inertial velocity matrices are calculated, for this we leave it expressed as can be seen in the \autoref{eq:MatrizInercial} and \autoref{eq:MatrizVelocidadAngular}.
The angular velocity and inertial matrices are calculated, can be seen in the \autoref{eq:MatrizInercial} and \autoref{eq:MatrizVelocidadAngular}.

    \begin{equation}
        I_i=\begin{bmatrix}I_{xxi}&I_{xyi}&I_{xzi}\\I_{yxi}&I_{yyi}&I_{yzi}\\I_{zxi}&I_{zyi}&I_{zzi}\end{bmatrix}
        \label{eq:MatrizInercial}
    \end{equation}

    \begin{equation}
        \begin{matrix}
    \omega_A=\begin{bmatrix}\dot{\theta}_1\\0\\0\end{bmatrix} & \omega_B=\begin{bmatrix}\dot{\theta}_1 +\dot{\theta}_2\\0\\0\end{bmatrix} & \omega_C=\begin{bmatrix}\dot{\theta}_1 + \dot{\theta}_2\\\dot{\theta}_3\\0 \end{bmatrix} 
    \end{matrix}
    \label{eq:MatrizVelocidadAngular}
    \end{equation}
    
    \[\begin{matrix}
     \omega_D=\begin{bmatrix}\dot{\theta}_1 + \dot{\theta}_2 \\\dot{\theta}_3 + \dot{\theta}_4\\0 \end{bmatrix} & \omega_E=\begin{bmatrix}\dot{\theta}_1 + \dot{\theta}_2 + \dot{\theta}_5\\\dot{\theta}_3 + \dot{\theta}_4\\0 \end{bmatrix} 
    \end{matrix}\]
    
    \[\omega_F=\begin{bmatrix}\dot{\theta}_1 + \dot{\theta}_2 + \dot{\theta}_5 + \dot{\theta}_6\\ \dot{\theta}_3  + \dot{\theta}_4\\0 \end{bmatrix}\]

\begin{itemize}
    \item Kinetic energy
    
% We proceed to calculate the total kinetic energy described below. Where the sub-indice $i$ indicate the potential energy on each body ($A, B, C, D, E$ and $F$) . 
Total kinetic energy is calculated in \autoref{eq:EnergiaCinetica}. Where the sub-indice $i$ indicate the kinetic energy on each body ($A, B, C, D, E$ and $F$).
    
\begin{equation}
        T_{total} = \sum_{i=1}^{6}{T_i},
        \label{eq:EnergiaCinetica}
\end{equation}

The kinetic energy of the bodies can be written in a generalized form (\autoref{eq:EnergiaCineticaGeneral}):

\begin{equation}
    T_i = \frac{1}{2} m_i \dot{r}_i^2 + \frac{1}{2} \omega_i^T I_{i} \omega_i
    \label{eq:EnergiaCineticaGeneral}
\end{equation} 

    \item Potential energy
    
The total potential energy is calculated in \autoref{eq:EnergiaPotencial}

\begin{equation}
    V_{total}=\sum_{i=1}^{6}{V_i}
    \label{eq:EnergiaPotencial}
\end{equation}

Potential energy of the bodies can be written in a generalized form (\autoref{eq:EnergiaPotencialGeneral}):

\begin{equation}
    V_i =  m_i \cdot g \cdot h_i 
    \label{eq:EnergiaPotencialGeneral}
\end{equation}
    
Where $m_i$ is the mass of each body, $g$ is the gravity and $h_i$ is the height of each body. 
    
\end{itemize}
\vspace{2mm}
% With the total kinetic and potential energies, we proceed to calculate the Lagrangian, from the \autoref{eq:Lagrangiano}. 
With the total kinetic and potential energies are proceed to calculate the Lagrangian (\autoref{eq:Lagrangiano}). 

\subsection{Lagrange differential equation}

The Lagrange differential equation is described in \autoref{eq:EDL}. 
    
\begin{equation}
     \frac{\mathrm{d} }{\mathrm{d} t}     \left( \frac{\partial \mathcal{L} }{\partial \dot{q}}\right)-
      \frac{\partial \mathcal{L} }{\partial q} =Q_{q}
      \label{eq:EDL}
\end{equation}

$\partial \mathcal{L}/\partial \dot{q}$ $\Longrightarrow$ Partial derivative of the Lagrangian with respect to the partial derivative of velocity.

$\partial \mathcal{L}/ \partial q$ $\Longrightarrow$ Partial derivative of the Lagrangian with respect to the partial derivative of position.

% The generalized forces are given as follows:
The generalized forces are:
    
\[Q_q = \tau - \beta\cdot \dot q\]

Where $\tau$ is the torque generated from the gear motor output, $\beta$ the viscous damping coefficient and $\dot q$ the velocity vector.

Motor torque is defined as:
\begin{equation}
    \tau = K_r\cdot k_{\phi_m} \cdot i_a
\end{equation}

% The input voltage is set to the \autoref{eq:VoltajedeEntradaMotor}.
The input voltage is described in \autoref{eq:VoltajedeEntradaMotor}.
    
\begin{equation}
V_a = R_a i_a + L_a \frac{di_a}{dt} + e_a
    \label{eq:VoltajedeEntradaMotor}
\end{equation}
\[V_a = R_a i_a + L_a \frac{di_a}{dt} + k_{\phi_m} \dot \theta_m\]
    
The electromotive force ($e_a$) is proportional to the angular velocity of the motor ($\theta_m$), $i_a$ is the armature current, $R_a$ is the armature resistance, $L_a$ is armature inductance, $di_a / dt$ is the voltage across the armature inductance, $k_{\tau}$ is the torque constant and $k_{\phi_m}$ is the against-electromotive constant.
% \begin{equation}
%     V_a = R_a i_a + L_a \frac{di_a}{dt} + k_{\phi_m} \dot \theta_m
% \end{equation}

Finally, the Euler-Lagrange equation is described in the \autoref{eq:Euler-Lagrange}.
    
\begin{equation}
     D(q)\ddot{q} + C(q,\dot{q})\dot{q} + G(q) =\tau
     \label{eq:Euler-Lagrange}
\end{equation}

\begin{center}
$D(q)$ $\Longrightarrow$ Matrix off mass and inertia

$C(q,\dot q)$ $\Longrightarrow$ Coriolis and centripetal effects

$G(q)$ $\Longrightarrow$ Stifness and gravitationals effects

$\tau$ $\Longrightarrow$ Input  
\end{center}

\subsection{Simulation of the dynamical model}

% En la \autoref{fig:EstadodeReposo}, se observa la simualción de modelo dinámico en estado de reposo, se visualiza con particularidad que Theta 1 ($\theta_1$) llega aproximadamenta a 90°, debido a que se simula como si de callera al suelo y rebotara hasta quedarse quieto, tambien se observa que el posición de Theta 3 ($\theta_3$) cae a -35° que equivale a 0.6 radianes aproximadamenta, esto significa que dado que la cadera queda un poco inclinada con respecto al suelo, su posición final en estado de reposo es inclinada. 

\begin{figure}[ht]
	\centering
		\includegraphics[trim=0 0 0 0,clip,width=0.48\textwidth]{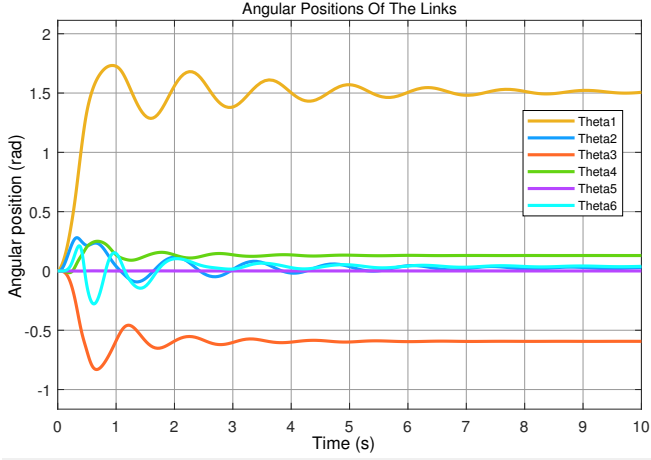}
		\caption{Response of the mathematical model to zero input}
	\label{fig:EstadodeReposo}
\end{figure}

% In the \autoref{fig:EstadodeReposo}, it is observed the simulation of dynamic model in resting state, it is visualized with particularity that Theta 1 ($\theta_1$) reaches approximately 90°, because it is simulated as if it fell to the ground and bounced to stand still.
In the \autoref{fig:EstadodeReposo}, it is observed the simulation of dynamic model to zero input, it is visualized with particularity that Theta 1 ($\theta_1$) reaches approximately 90°, because it is simulated as if it fell to the ground and bounced to stand still.

It is also observed that the position of Theta 3 ($\theta_3$) falls to -35° which is equivalent to 0.6 radians approximately, this means that since the hip is slightly tilted with respect to the ground, its final position at rest is tilted. 

%-9.80665

\subsection{Proposed control}

% A non-linear control designed for the three-dimensional movement of the BLUE robot is proposed, where there are six control inputs which drive the system. The proposed control scheme is made up of the matrix $ D $, $ C $ and the vector $ G $. (\autoref{eq:control1})

It is proposed a non-linear control designed for the three-dimensional movement of the BLUE robot, where there are six control inputs which drive the system. The proposed control scheme has matrix $ D $, $ C $ and the vector $ G $.

% Considering the Euler-Lagrange model and solving for the higher-order derivative, the following expression is obtained:
The higher-order derivative is solved as of Euler-Lagrange model and obtained \autoref{eq:control1}.
 \begin{equation}
     \ddot{q}=D^{-1}(q)*[-C(q,\dot{q})*\dot{q}-G(q)]+D^{-1}(q)*\tau
    \label{eq:control1}
 \end{equation}

%  The proposed control law for each signal of each motor is described as follows: 
The proposed control law for signal of each gear motor is described as:
\[ \tau=D(q)*[D^{-1}(q)*(C(q,\dot{q})+G(q))+\mu]\]
 
 Where $\mu$ and the error $e$ it's defined to: 
\[ \mu=\ddot{q}_{d}-k_{p}e-k_{d}\dot{e} \]
\[ e=q-q_{d}\]

$\dot{e}$ $\Longrightarrow$ Velocity error ($\dot{q}-\dot{q}_d$).

$\ddot{q}_{d}$ $\Longrightarrow$ The desired acceleration.

$\dot{q}_d$ $\Longrightarrow$ The desired velocity.

$q_d$ $\Longrightarrow$ the desired position.

$\dot{q}$ $\Longrightarrow$ Measured velocity.

$q$ $\Longrightarrow$ Measured position.

% A stabilization time of $ t_ {s} = 0.5s $ was defined with a dominant pole at $ (s-8) $ and a polo six times further away at $ (s-40) $. In this way the gains of the control system are obtained, which are $ k_p = 320, k_d = 48 $. Tracking of the desired trajectory is checked, making use of the proposed control system. The pursuit of the desired trajectories can be evidenced in \autoref{fig:Seguimiento}

\begin{figure}[ht]
	\centering
		\includegraphics[trim=0 0 0 0,clip,width=0.48\textwidth]{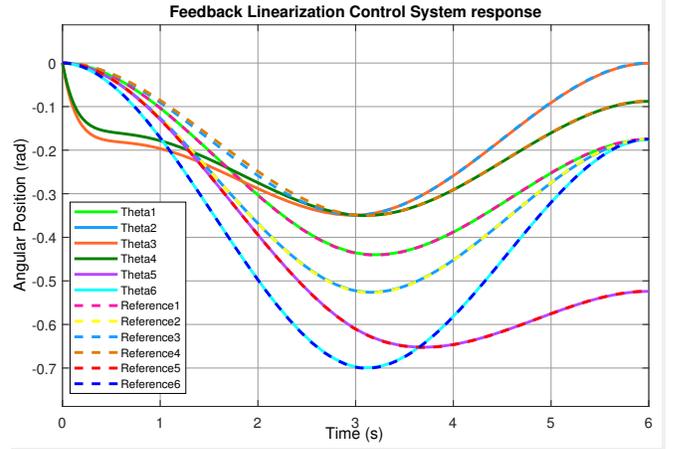}
		\caption{Feedback Linearization Control System response.}
	\label{fig:Seguimiento}
\end{figure}

A stabilization time of $ t_ {s} = 0.5s $ was defined with a dominant pole at $ (s-8) $ and a polo five times faster $ (s-40) $. The gains of the control system are $ k_p = 320$ and $k_d = 48 $. Tracking of the desired trajectory is checked, making use of the proposed control system. Follow up of the desired trajectories can be evidenced in \autoref{fig:Seguimiento}.

\section{Experimental test}

\begin{figure}[ht]
	\centering
		\includegraphics[trim=0 0 0 0,clip,width=0.43\textwidth]{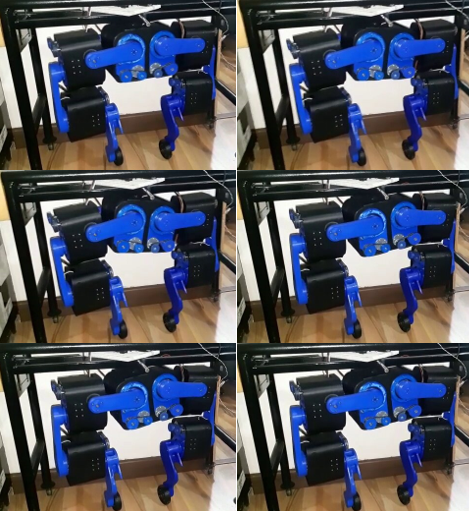}
		\caption{Hip movement sequence}
	\label{fig:Exp1}
\end{figure}

The repetitive movement of each of the joints was developed to verify compliance in the mobility of each BLUE joint, as a support, video capture was made, which is observed in \autoref{fig:Exp1} fragmented for its correct visualization. Within these practices standard resources such as an Arduino Nano and L298N module.

\begin{figure}[ht]
	\centering
		\includegraphics[trim=0 0 0 0,clip,width=0.43\textwidth]{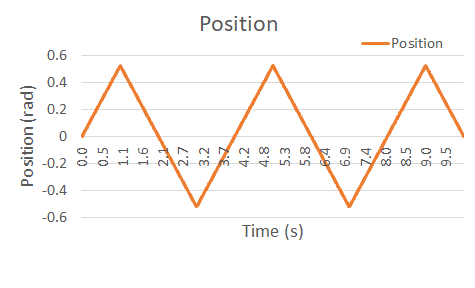}
		\caption{Hip encoder reading.}
	\label{fig:ENC1}
\end{figure}

\autoref{fig:ENC1} shows the encoder reading of the hip actuator where the angle varies from $30$ to $-30$ degrees.

\newpage
\section{Conclusions}
A 3D bipedal robot was designed and fabricated using a mechatronics designing approach. To reduce the impact effects on the support-foot exchange event, a flexible serial actuator was developed. Additionally, a reduction of the mass was performed with a generative design approach, which uses topology optimization to define the shape of the robot's shins. These mechanisms and parts were successfully fabricated and implemented using 3D-printed parts. The bipedal robot was evaluated numerically and a prototype was fabricated to validate the design.     
%It was concluded that the optimal material for the development of the research was the use of PLA in 3D printing due to its easy handling and the mechanical properties offered with a 90 percent filling due to the fact that, as observed in the \autoref{tab:VelocidadesyTorquesIO} parts, it offered quite a range of operation in terms of applied torques, radial forces, and tangential forces.
%
%Regarding the design of the shin of the BLUE robot, it was found that the use of the topological optimization technique was valid for the implementation in the final prototype as observed in the \autoref{fig:Shindesignopt}. This, due to the possibility of resisting normal and shear stresses in a similar way to the initially proposed piece, in the same way the optimization allowed savings in the use of material necessary for its development without affecting the properties of the mechanical leg at any time.
%

The preliminary results from the design of a nonlinear control with feedback linearization show acceptable performance in tracking trajectories.

\bibliographystyle{ieeetr}
\bibliography{biblio.bib}

\end{document}